\documentclass[letterpaper, 10 pt, conference]{ieeeconf}  % Comment this line out if you need a4paper

\IEEEoverridecommandlockouts  
\usepackage{graphicx}
\usepackage{amsmath}
\usepackage{amssymb}
\usepackage{booktabs}

\usepackage{pifont}
\let\labelindent\relax
\usepackage[inline]{enumitem}
\newcommand{\cmark}{\ding{51}}%
\newcommand{\xmark}{\ding{55}}%
\usepackage{url}
\usepackage{graphicx}

\usepackage{booktabs} 
\usepackage{floatrow}
\usepackage{subcaption}
\newfloatcommand{capbtabbox}{table}[][3\FBwidth]
\usepackage[para]{threeparttablex}
\usepackage[pagebackref=true,breaklinks=true,colorlinks,bookmarks=false]{hyperref}

\def\BibTeX{{\rm B\kern-.05em{\sc i\kern-.025em b}\kern-.08em
    T\kern-.1667em\lower.7ex\hbox{E}\kern-.125emX}}

\title{\LARGE \bf

Towards Inclusive HRI: Using Sim2Real to Address Underrepresentation in Emotion Expression Recognition}

\author{Saba Akhyani, Mehryar Abbasi, Mo Chen, and Angelica Lim% <-this % stops a space
\thanks{All authors are associated with Simon Fraser University, Burnaby, Canada {\tt\small \{sakhyani,mabbasib,mochen,angelica\}@sfu.ca}. This work was supported by the Huawei-SFU Visual Computing Joint Lab and the Natural Sciences and Engineering Research Council of Canada (NSERC), RGPIN/06908-2019.}%
}

\begin{document}

\maketitle
\thispagestyle{empty}
\pagestyle{empty}

%%%%%%%%%%%%%%%%%%%%%%%%%%%%%%%%%%%%%%%%%%%%%%%%%%%%%%%%%%%%%%%%%%%%%%%%%%%%%%%%
\begin{abstract}

Robots and artificial agents that interact with humans should be able to do so without bias and inequity, but facial perception systems have notoriously been found to work more poorly for certain groups of people than others. In our work, we aim to build a system that can perceive humans in a more transparent and inclusive manner. Specifically, we focus on dynamic expressions on the human face, which are difficult to collect for a broad set of people due to privacy concerns and the fact that faces are inherently identifiable. Furthermore, datasets collected from the Internet are not necessarily representative of the general population.
We address this problem by offering a Sim2Real approach in which we use a suite of 3D simulated human models that enables us to create an auditable synthetic dataset covering
1) underrepresented facial expressions, outside of the six basic emotions, such as confusion;
2) ethnic or gender minority groups; and 3) a wide range of viewing angles that a robot may encounter a human in the real world. By augmenting a small dynamic emotional expression dataset containing 123 samples with a synthetic dataset containing 4536 samples, we achieved an improvement in accuracy of 15\% on our own dataset and 11\% on an external benchmark dataset, compared to the performance of the same model architecture without synthetic training data. We also show that this additional step improves accuracy specifically for racial minorities when the architecture's feature extraction weights are trained from scratch.

\end{abstract}

%%%%%%%%%%%%%%%%%%%%%%%%%%%%%%%%%%%%%%%%%%%%%%%%%%%%%%%%%%%%%%%%%%%%%%%%%%%%%%%%
\section{Introduction}
\label{sec:intro}

There has been an increasing interest in using robots in everyday social environments like hospitals, retail stores, and homes. As a result, it has become essential for robots to communicate and interact socially for various human-robot interaction (HRI) applications and understand people's nonverbal expressions. For instance, imagine a restaurant service robot seeing a look of confusion wash over your face as you look at the menu. It detects your hesitation and proactively offers assistance: ``Do you have any questions I can answer?" This simple scenario plays out between humans each day all over the world, but robots are still far from capable of performing this kind of proactive assistance in a robust manner. Several major challenges prevent such systems from being deployed in the real world.

Firstly, a recent review has argued that state-of-the-art facial emotion classifiers cannot be applied effectively to human emotion analysis in the wild~\cite{LFbarret}.
One underlying reason is that in HRI, as stated in a review by~\cite{McColl2015}, ``there are a wide range of possible affective levels expressed by people in human-robot interaction that the robot needs to understand in order to participate in a natural bi-directional social interaction with humans." In other words, real-world interactions comprise a rich and subtle set of expressions, while most datasets focus on collecting the prototypical set of emotions~\cite{LFbarret} of happiness, sadness, anger, surprise, fear, disgust, and neutral~\cite{jangid2019video,tang2013deep,yu2015image,kim2016fusing}, 
Specifically, to the best of our knowledge, there is no public video dataset containing confusion~\cite{Yasser2021} nor any comparison benchmark~\cite{Hucko2020} for these dynamic facial expressions~\cite{Jack2017,dornaika2007efficient}. Strategies are needed to effectively create video datasets for the many underrepresented emotional expression categories that \emph{actually} occur in the wild with robots, such as the 28 social signals identified in Saheb Jam et al.~\cite{SahebJam2021}, including confused, worried, skeptical, and so on. Indeed, prototypical sadness or fear were not seen in these real-world human-robot interactions, and we do not focus on these emotions in this paper.

Secondly, social robots should also have the ability to evaluate human affective expressions fairly, without discriminating against underrepresented groups. A recent survey on automatic multi-modal emotion recognition in the wild shows that inclusivity of all ethnicities remains a challenge in facial expression recognition systems and should be further investigated~\cite{Sharma2020}. According to~\cite{Rhue2018,pmlr-v81-buolamwini18a}, racial bias is apparent in current machine learning methods in general, especially those involving the face. %In one study,~\cite{pmlr-v81-buolamwini18a} identified that datasets are generally composed of lighter-skinned subjects. They also found out that the commercial face gender classification systems perform better on male and light faces, and are least accurate on dark-skinned females. In another study, Rhue~\cite{Rhue2018} used a dataset of basketball player photos and found racial biases in Face++ and Microsoft's Face API. It was more likely for Black men to be tagged with a negative emotion than white men. 
One major cause of this bias is that major facial expression recognition (FER) datasets are underrepresentative of genders and non-Caucasian backgrounds~\cite{tommasi2017deeper}. As faces are inherently identifiable, ethical and privacy concerns arise when requiring real humans to provide their data~\cite{Jordan2015}, yet anonymizing the face can remove important facial features. Data collection to reduce bias in face-related algorithms is, therefore, a major ethical challenge.
   
Finally, mobile robots can potentially view humans from various viewing angles and in varied lighting conditions. Collecting and labeling large amounts of naturalistic videos for facial expression recognition is challenging due to several reasons. First, creating datasets for spontaneous (rather than posed) user affective states is very time-consuming~\cite{sims_neural_2020}. Additionally, facial emotional expressions are difficult to label due to the subjective nature of annotation, compared to those in domains where deep learning has been most successful, such as in object recognition. Thus, data augmentation techniques are expected to be particularly useful to help facial expression recognition succeed in the wild. 

% The other critical problem is that after gathering real datasets, biases become apparent in the composition of genders and races~\cite{tommasi2017deeper}. Therefore, multiple studies have utilized the generation of synthetic data to create a large dataset~\cite{abbasNejad, Han}, mitigating bias and the difficulty in collecting real data~\cite{Kortylewski2018TrainingDF}. While studies have generated synthetic data to augment underrepresented samples for the face related tasks~\cite{KortyAnalzye2019}~\cite{Ngxande2020}, these studies are for tasks other than facial emotion recognition. Therefore, there is still a gap on inclusive datasets of dynamic facial emotion expressions for underrepresented ethnicities.

In this paper, we employ a simulation to reality (Sim2Real) approach to address the previously mentioned challenges. Sim2Real approaches have performed well in different domains such as hand tracking~\cite{GANeratedHands_CVPR2018} and text detection~\cite{gupta2016synthetic}. For face recognition and facial feature detection, studies have attempted to address some of these problems by creating static, synthetic image  datasets~\cite{gerig2018morphable,abbasNejad,Han2018ImprovingFD,kortylewski2018empirically}. We extend these approaches into the video emotion recognition domain by \begin{enumerate*}[label=\roman*)]\item identifying facial movements and social signals of the desired dynamic emotional expressions from real data, \item converting the identified social signals into renderable animations, \item generating virtual human models of various and specific ethnicities, \item rendering the animations into videos of all virtual human models from multiple viewing angles, \item pretraining a dynamic emotion classifier model on the synthetic videos, \item retraining and testing the model on the real dataset.
\end{enumerate*}

\begin{figure}[!t]
\centering
          \includegraphics[width=0.7\textwidth]{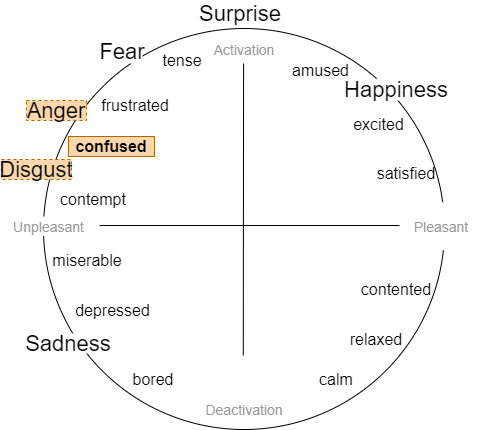}
         \caption{Circumplex Model of Affect adapted from \cite{Russell1980}.}\label{fig:diag}
     %\hfill
\end{figure}

We generate an inclusive synthetic facial expression dataset from virtual humans that specifically incorporates new expressions outside of the prototypical set of six basic emotions, underrepresented ethnicities, and varied camera angles. As a proof-of-concept, we detect the dynamic social signal of confusion. In order to evaluate the model, anger and disgust were chosen from the 6 basic emotions as challenging  emotion expression confounders due to their similarity to confusion. Characteristics of these three emotions overlap, and expressions can be easily mistaken for one another, as illustrated in the Circumplex Model of Affect (CMA) shown in Fig.~\ref{fig:diag}. The circumplex model is a graphical representation of affective states on the 2-D plane of arousal (vertical) and valence (horizontal).

Our contributions are as follows:
 \begin{enumerate}
% CVPR reviewer said this claim is not right
%\item To the best of our knowledge, this paper proposes for the first time a \textbf{Sim2Real} approach for facial expression recognition, capable of confident predictions with limited video data
 \item We propose the first video dataset of the understudied emotional expression of confusion (as well as nearby social signals ``disgust" and ``anger" as shown in the circumplex model in Fig.~\ref{fig:diag}). %To the best of our knowledge and as mentioned in Yasser et al.~\cite{Yasser2021}, there is no public dataset available for confusion detection. 
 We first gather a real dataset, then augment this limited data using Sim2Real to produce a much larger synthetic dataset. Our dataset also addresses the critical issue of racial bias, which is apparent in existing real-world data, by comprising faces of underrepresented ethnicities, including Black, Asian and Hispanic individuals.

 \item We explore the effect of adding synthetic data on improving fairness using a CNN+Time Series Classification (TSC) network architecture. Our experiments demonstrate that: 1) training on a combination of real-world data and a randomly selected portion of synthetic data (changing every epoch) achieves the highest performance and 2) fine-tuning on a pre-trained CNN with \emph{unfrozen face feature extraction weights} decreases racial bias.
\end{enumerate}

The Sim2Real approach would allow us to create even larger synthetic datasets in future, because of the flexibility to be applied to any desired emotion or ethnicity. This is feasible due to the fact that facial movements (action units) associated with any emotion can be extracted either automatically using OpenFace~\cite{amos2016openface} or by manual observation. Additionally, the wide modification range of the MakeHuman toolkit~\cite{bastioni2008idea} allows for the generation of several human models to incorporate other ethnicities.

\section{Methodology}
The methodology behind our Sim2Real approach is explained in this section. An overview of our dataset generation and preparation, as well as an overview of our deep-learning-based dynamic facial expression recognition model, is provided. The synthetic data pretraining step of the Sim2real approach is explored in Section~\ref{sec:exp}.
\subsection{Dataset Generation}
In this section, we describe the collection of in-the-wild emotionally expressive videos and the generation of synthetic videos using a suite of simulated humans. We focus this study on confusion, a dynamic social signal which is underrepresented in datasets and lacks examples on the web~\cite{Hucko2020}, yet is common in HRI~\cite{SahebJam2021}.

\subsubsection{Collection of in-the-wild confusion, anger, and disgust videos}
% Confusion has specific characteristics, such as eyelid tightening and/or frowning, which can be mistaken for social signals of anger and disgust. We, therefore, focus on making a dataset for these three emotions.
Confusion is an affective state conveyed through varied and multiple expressions, as are both anger and disgust. Some of these expressions (such as frowning) are common between the three, resulting in some expressions being easily mistaken as another emotion, possibly due to these three emotions appearing very close together in the CMA~\cite{Russell1980}. We therefore focus on making a dataset for these three emotions.

\begin{figure}[t]
\centering
          \includegraphics[width=0.7\textwidth]{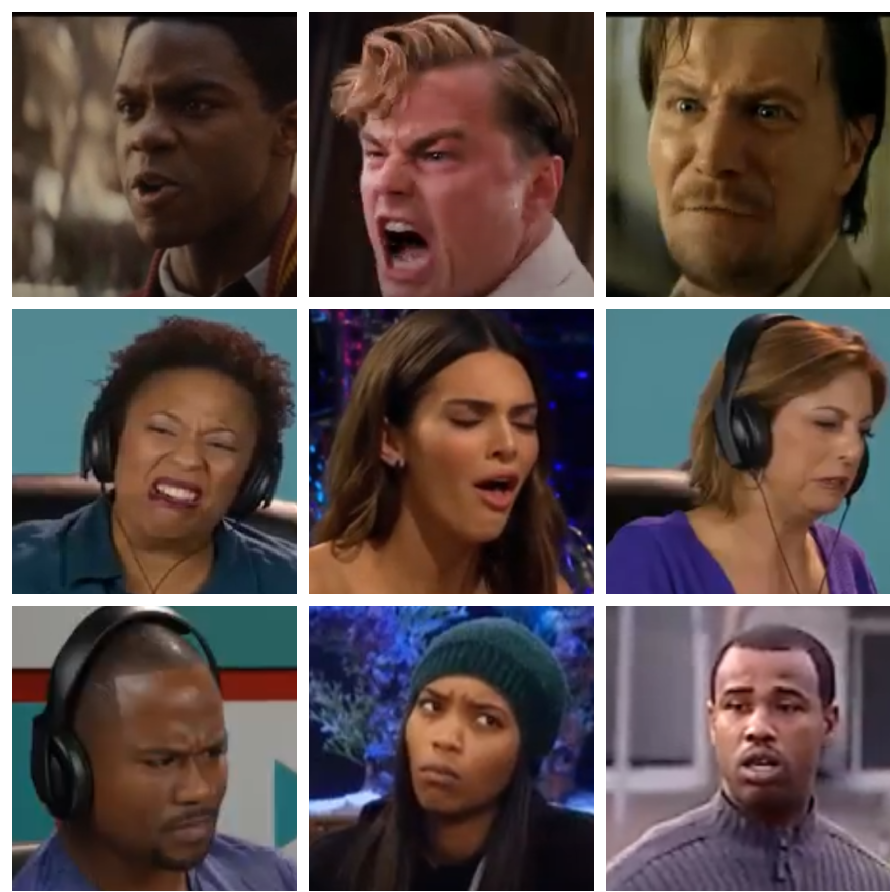}
         \caption{Examples of real videos: the first row is for angry, second for disgusted, and third row for confused expressions}\label{fig:re}
     \hfill
\end{figure}

We collected short video clips (1-3s) from YouTube.com and Giphy.com using search tags such as ``angry", ``confused", and ``disgust" reactions to gather human facial expressions of our desired social signals, as shown in  Fig.~\ref{fig:re}. This search resulted in 153 clips. Each video was then labeled for the conveying facial expression, by two annotators identifying with Canadian culture (inter-rater agreement kappa score=.88), and low confidence videos were discarded. 

We created a multi-ethnicity dataset of real human videos expressing the three social signals of confusion (41 videos), anger (41 videos), and disgust (41 videos). The final dataset contains 123 videos, of which 26 are of non-Caucasian individuals. 

\subsubsection{Generation of augmented dynamic social signal video dataset}

The task of creating desired social signals videos is made possible using the MakeHuman toolkit~\cite{bastioni2008idea} and the FACSHuman plugin~\cite{gilbert2018facshuman}. MakeHuman toolkit is an open-source and free 3D computer graphics toolset designed for prototyping human-like models.
FACSHuman offers the possibility of manipulating the Action Units (AU) presented in the Facial Action Coding System (FACS)~\cite{ekman1997face} on the 3D models created in the MakeHuman software. This manipulation of AUs is a key component of our  Sim2Real process. FACSHuman enabled us to generate social signal animations that can be rendered into videos or frames on a selected human virtual model from any viewing angle. We created a script plugin that used this capability to render 4536 synthetic videos from the combination of 24 human virtual models, 21 social signals (facial movement animations), and 9 viewing angles.

\begin{figure}[!t]
\centering
          \includegraphics[width=0.6\textwidth]{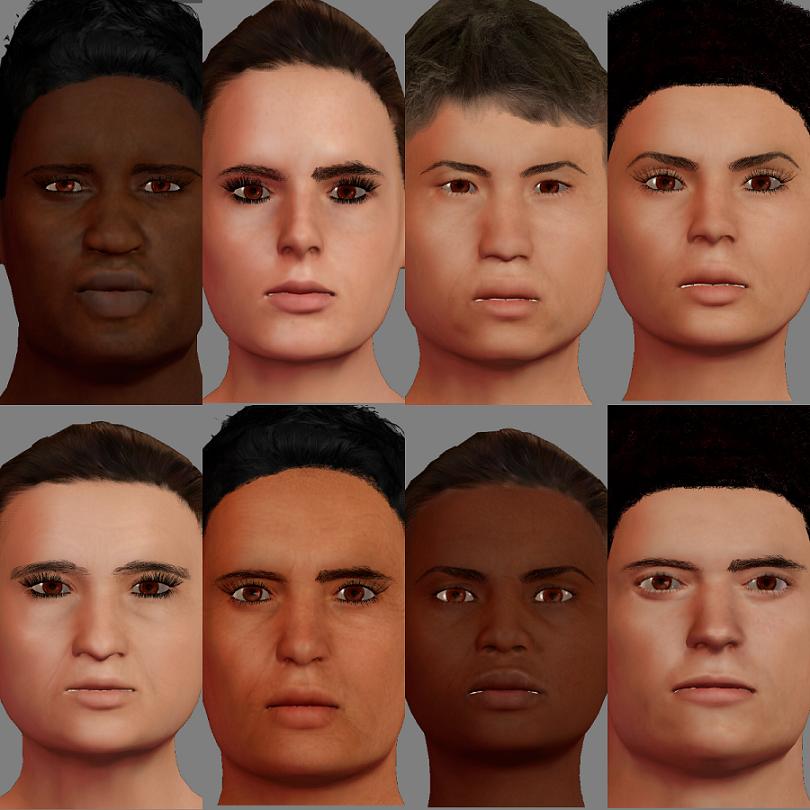}
         \caption{Sample of generated human-like models}\label{fig:mos}
     \hfill
\end{figure}

\textbf{Creation of a suite of simulated humans}: 
The overarching vision of this work is to create a large, auditable suite of human models to represent people from many different backgrounds. As a first step, we create 24 simulated human adult models balanced on gender and four different ethnicities (Caucasian, Black, Asian, and Hispanic). The MassProduce plugin within the MakeHuman application was then used to create several randomly generated human models of multiple ethnicities and different ages and skin colors. Out of all those generated human-like models, we selected 24 models for our study based on the realism of action unit manipulation on the model (8 samples are shown in Fig.~\ref{fig:mos}). We chose to use 3D models as we hope to eventually use them in HRI simulators, to create expressive virtual humans with facial expressions. These models are provided on Github\footnote{\url{https://github.com/sabaak95/confusionDetection}} so that researchers can also import the 24 virtual humans with 21 dynamic expressions (7 social signals per emotion), to replay them in front of their virtual robot.
%Made the pic smaller for more visibility, and more balance %

\begin{figure}[!t]
 \centering
          \includegraphics[width=0.9\textwidth]{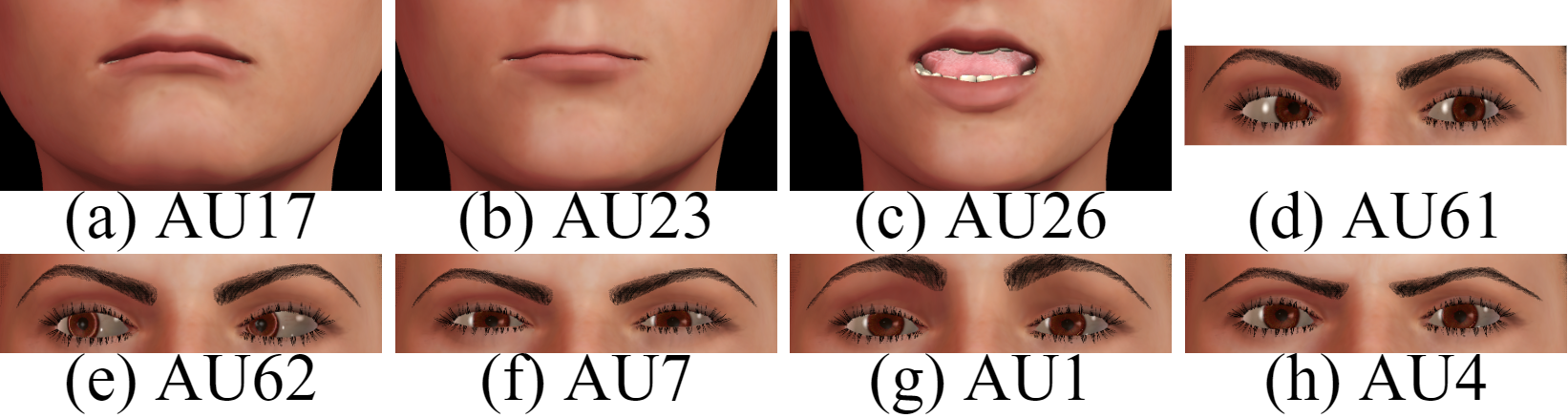}
  \caption{(a) Chin raiser, (b) Lip tightener , (c)  Jaw drop, 
    (d) Eyes left, (e) Eyes right, (f) Lid tightener, (g) Inner brow raiser, (h) Brow lowerer }\label{fig:au}%
\end{figure}

\textbf{Multiple social signals per emotion}: 
We used the FACSHuman software to create 21 different social signal animations, 7 for each emotional class. These animations convey multiple variations of social cues of confusion, anger, and disgust. These 21 social signals were manually animated over 25 frames and were created via inspection of the in-the-wild real human dataset. For example, Fig.~\ref{fig:au} shows the AUs that were used to create dynamic confusion social signals. Varied AU combinations and sequences were used to animate the 21 social signals, validated by an annotator with Canadian culture. An example is a side-eye movement confusion state made by a timed sequence of the following AUs: AU61, AU62, AU61. While future work should automatically perform the animation creation process from video data, the manual animation creation step in this study allows us to validate the Sim2Real portion given human-level feature extraction, enabling us to identify specific underlying social signals for each emotion (7 for each emotion), which are now available for use in our 3D models. The dataset is available for download.\footnote{ \url{https://www.rosielab.ca/datasets/confusion-in-the-wild}}

\begin{figure}[t]
         \centering
         \includegraphics[width=0.7\textwidth]{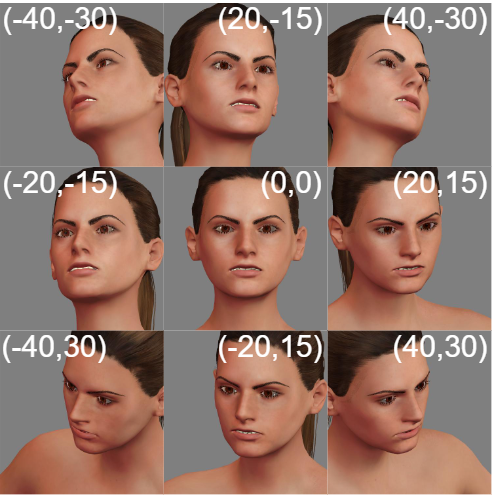}
         \caption{The 9 viewing angles used to generate our augmented dataset.}\label{fig:an}
\end{figure}

\textbf{Multiple viewing angles}:
As robots may view a human from varied angles, it is important that our generated dataset incorporate varied perspectives. Our dataset was therefore expanded by creating videos of the same facial gesture from 9 viewing angles, to make our network invariant to the face viewing angle, as shown in Fig.~\ref{fig:an}. The camera movement included horizontal rotations of $-40,-20,0,20,40$ degrees and vertical rotations of $-30,15,0,15,30$ degrees. The nine combinations of $(H_{rotation},V_{rotation})$=\{$(-40,-30)$, $(-20,-15)$, $(0,0)$, $(20,15)$, $(40,30)$, $(40,-30)$, $(20,-15)$, $(20,-15)$, $(40,-30)$\} were selected as our viewing angles in degrees, as shown in Fig.~\ref{fig:an}.
\subsection{Data preparation}
\begin{figure}[t]
  \centering
  \includegraphics[width=0.7\textwidth]{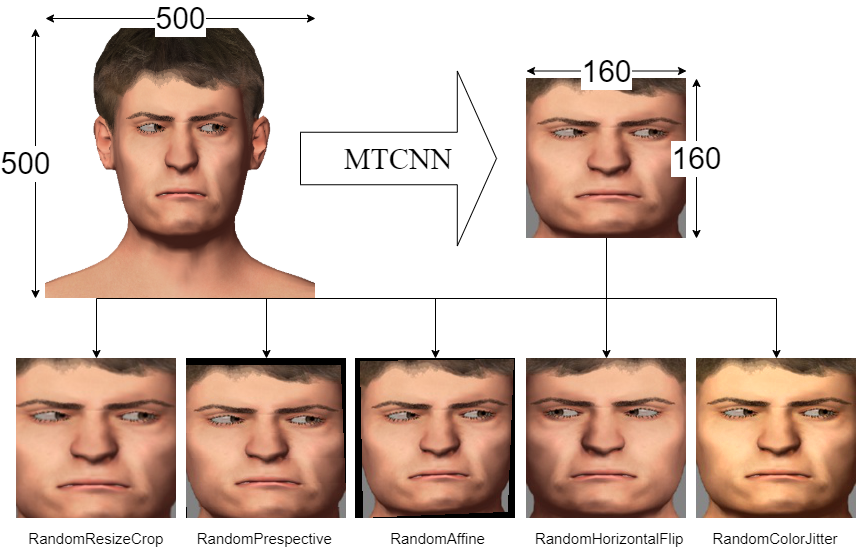}
  \caption{Image preprocessing and augmentation.}\label{fig:tr}
\end{figure}

In order to refine the data and remove any unimportant or unrelated information in the images, we used Multi-task CNN (MTCNN) to detect and crop the faces before feeding frames to our network~\cite{zhang2016joint} and resized images to $160\times160$. Additional transforms were also applied to the images randomly on each epoch, including cropping, perspective, affine transform, horizontal flip, and color transforms (shown in Fig.~\ref{fig:tr}).%
We ensured that the same transformations were applied to all the frames from the same video.

%%%%% below part from sim2real fg v1, new added data, which I find useful ++ some new data
%%%%

\subsection{Model Architecture}
We developed a basic framework for our video classification problem to test the Sim2Real strategy. Existing work can extract valuable frame-based facial features from a face image, such as FaceNet~\cite{schroff_facenet_2015} and OpenFace~\cite{amos2016openface}; one of these models can be used to first extract each frames' facial features. After frame-based feature extraction, we model the problem as Time Series Classification. Sections~\ref{FFE},~\ref{TSC} are dedicated for further exposition on our selections for this architecture. 

\subsubsection{Facial Feature Extraction Network}\label{FFE}
We used the pre-trained FaceNet~\cite{schroff_facenet_2015} architecture as our facial feature extractor. FaceNet uses an InceptionResnetV1 architecture trained on the VGGFace2 dataset. Each video is given to FaceNet frame by frame, and the output feature arrays are concatenated together across the time dimension to create a multivariate time series array. 
\subsubsection{Time Series Classifier Network}\label{TSC}
Further processing of this output requires a time series classification algorithm. K-Nearest Neighbor (KNN) algorithm with Dynamic Time Warping (DTW)~\cite{DTW} metric is one of the earliest techniques for this task that is still used, specially when working with relatively small datasets. Many machine learning algorithms have also been applied to this problem, such as ResNet and FCN~\cite{Resnetime}. However, we opted to use InceptionTime~\cite{inceptiontime} as our classifier because it has proven to be a versatile and promising machine learning solution for many Time Series Classification tasks, based on its performance results on the UCR~\cite{UCR} benchmark collection datasets.
\subsubsection{Proposed and Baseline Architectures}
\begin{figure}[t]
  \centering
  \includegraphics[width=0.7\textwidth]{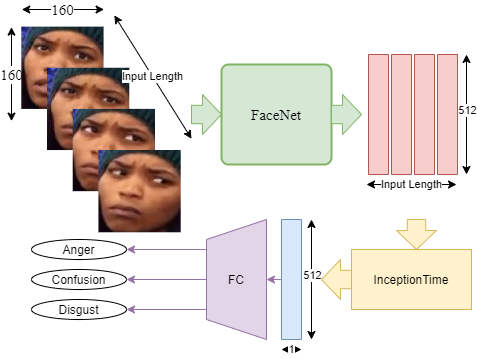}
  \caption{Model architecture for FaceNet+InceptionTime}\label{fig:mod}
\end{figure}

Our proposed DNN structure is the combination of FaceNet~\cite{schroff_facenet_2015} and InceptionTime128~\cite{inceptiontime}. This structure is shown in Fig.~\ref{fig:mod}. This model is referred to here as FN+INC25 or FN+INC64. The two numbers of 25 and 64 indicate the required number of frames for the video input. Videos shorter than the indicated number are padded to the required input length, and the longer videos are cropped. This procedure is explained in Section~\ref{sec:exp}. We included two instances of these models in our experiments. In one instance, the FaceNet weights are frozen. In the main instance, the FaceNet weights are not frozen, and are tuned alongside the InceptionTime weights. We hypothesized that the addition of synthetic data would allow us to tailor the FaceNet weights in favor of this dataset.

The second DNN architecture included in our experiments is I3D~\cite{I3D}, an advanced video classification method that applies combined temporal and spatial processing using 3D convolutional layers. This addition enabled us to evaluate the problem using a model not already affected by previous facial information knowledge.

Finally, we included a KNN classifier architecture applicable to the small real dataset. This model uses FaceNet~\cite{schroff_facenet_2015} for frame facial feature extraction and a KNN with DTW~\cite{DTW} metric as the video classifier. We propose this baseline model, FN+KNN, for comparison.

\section{Experiments and  Results}\label{sec:exp}

In this section, we evaluate our network on the created real dataset. We performed several experiments varying the architecture, input length, and the use of only synthetic, synthetic plus real, or only real training data. We used 5-fold cross-validation on the \textit{real dataset} to compare the performance of different approaches, with one fold consisting primarily of expressions by non-Caucasian individuals.

We explored three training strategies for our experiments. In the first strategy, the algorithms are only trained on the small real dataset. The baseline KNN model was tested under this strategy. In the second strategy, the networks are first trained on synthetic data, then fine-tuned on the real dataset. The second strategy was developed to add and assess the addition of synthetic data. Third, the  strategy was to combine the real training data set with the synthetic dataset and pass them to the network alongside each other. This strategy was designed to explore if a higher performance could be achieved by creating ratioed synthetic and real data training. Instead of combining the whole synthetic dataset with the real data, we trained the network with one-fourth of the synthetic dataset. In the next test, the ratio of the synthetic dataset was set to half. Finally, in the last test, the ratio of the synthetic dataset was set to one, meaning the whole synthetic dataset was included.

One important factor in our training and testing processes is setting the input video length to a fixed number of frames $L$. Input videos shorter than the set length were looped until they reached length $L$. The way we dealt with longer videos differed depending on the training phase. In the training phase, we randomly selected $L$ consecutive frames from the lengthier videos. For a video with $N$ frames, frames $n$ to $n+L-1$ are cropped. The $n$ is selected randomly on each epoch between $0$ and $N-L$. However, in the testing phase, we only selected the middle $L$ frames as the representative sequence in each video. In our experiment, we set $L$ to two values: 25 and 64. We selected 25 because the number of frames in our synthetic videos was 25. The choice of 64 was reliant on two factors. First, 64 was long enough to include a majority of the input video while small enough to keep computational cost and time consumption adequate. Second, the I3D network used in some of our experiments was designed for 64 frame inputs. In the following subsections, we elaborate on the latter two training strategies: \begin{enumerate*}[label=(\roman*)]
    \item fine-tuning the synthetic trained network on real data and combined synthetic.
    \item Combined synthetic and real data training.
\end{enumerate*}

\subsection{Fine-tuning the synthetic trained network on real data}

In this experiment, the model was first trained on the synthetic dataset alone. The simulated human models were randomly divided into two sets of 19 and 5 models. All of the generated videos using simulated human models in the larger set were used for training, and those in the smaller set were used for validation. The respective numbers for the videos in the training and validation data were 3591 and 945. The training was done over 20 epochs with the learning rate of $10^{-4}$ and the categorical cross-entropy loss function. The batch size was set to 8. After the training on synthetic data, we fine-tuned the model on the four selected training folds of the real dataset, over 50 epochs. The model is tested on the remaining single test fold. This operation is repeated 5 times each time a new fold is selected as the test fold. Learning rates and parameters were chosen empirically.\looseness=-1
\begin{table}[t]
\setlength{\tabcolsep}{3pt}
\renewcommand{\arraystretch}{0.05}
  \centering
  \caption{Performance comparison of all models}
   % Table generated by Excel2LaTeX from sheet 'ACC_F_RC2_9'
    \begin{threeparttable}
\begin{tabular}{l|cccccc}
\toprule
Network & Length & Syn\tnote{1} & \%Prc\tnote{2} & \%Rec\tnote{3} & \%Fs\tnote{4} & \%Acc\tnote{5} \\
\midrule
\midrule
FN(FZ\tnote{*} )+KNN & 25    & \xmark & 75    & 71    & 70    & 72 \\
\midrule
FN(FZ\tnote{*} )+KNN & 64    & \xmark & 79    & 74    & 73    & 74 \\
\midrule
FN(FZ\tnote{*} )+KNN & 283   & \xmark & 77    & 71    & 69    & 71 \\
\midrule
\textbf{FN(FZ\tnote{*} )+IN25} & 25    & \xmark & 76    & 73    & 73    & 73 \\
\midrule
\textbf{FN(FZ\tnote{*} )+IN25} & 25    & \cmark & 82    & 79    & 79    & 80 \\
\midrule
\textbf{FN(FZ\tnote{*} )+IN64} & 64    & \xmark & 86    & 71    & 71    & 72 \\
\midrule
\textbf{FN(FZ\tnote{*} )+IN64} & 64    & \cmark & 85    & 82    & 82    & 82 \\
\midrule
\textbf{FN+IN25} & 25    & \xmark & 78    & 74    & 74    & 74 \\
\midrule
\textbf{FN+IN25} & 25    & \cmark & \textbf{90} & \textbf{89} & \textbf{89} & \textbf{89} \\
\midrule
\textbf{FN+IN64} & 64    & \xmark & 81    & 77    & 76    & 77 \\
\midrule
\textbf{FN+IN64} & 64    & \cmark & 88    & 87    & 87    & 87 \\
\midrule
I3D   & 64    & \xmark & 66    & 63    & 60    & 66 \\
\midrule
I3D   & 64    & \cmark & 85    & 83    & 83    & 83 \\
\bottomrule
\end{tabular}%
\begin{tablenotes}
  %\item[*] Single  Image Input
  \item[*] Frozen FaceNet weights
  \item[1] Synthetic-Data \item[2]  Precision \item[3]  Recall \item[4]  F-score \item[5]  Accuracy 
  \end{tablenotes}
  \end{threeparttable}

  \label{tb:1}%
\end{table}%

 The results averaged over all 5 runs for these experiments are shown in Table~\ref{tb:1}. FZ specifies the instances where FaceNet weights were frozen to treat FaceNet purely as a feature extraction network, with weights of the rest of the network updated during training. Table \ref{tb:1} also includes experiments in which the synthetic data training step was skipped to highlight its effect. Additionally, we compared our methods with the baseline FN+KNN classifier applied only to the real data. 
\begin{table}[t]
\setlength{\tabcolsep}{5pt}
\renewcommand{\arraystretch}{0.05}
  \centering
  \caption{Effect of the addition of synthetic data and weight unfreezing on the non-Caucasian Fold}
    \begin{tabular}{lcc||cc}
    \toprule
    \multicolumn{1}{c}{NET} & Len   & Synth & \%Acc   & \%Acc Increase \\
    \midrule
    \midrule
    FN(FZ)+IN25 & 25    & \xmark & 72    & Base \\
    \midrule
    FN(FZ)+IN25 & 25    & \cmark & 80    & +8 \\
    \midrule
    FN+IN25 & 25    & \xmark & 80    & +8 \\
    \midrule
    FN+IN25 & 25    & \cmark & 88    & +16 \\
    \midrule
    FN(FZ)+IN64 & 64    & \xmark & 64    & Base \\
    \midrule
    FN(FZ)+IN64 & 64    & \cmark & 72    & +8 \\
    \midrule
    FN+IN64 & 64    & \xmark & 80    & +16 \\
    \midrule
    FN+IN64 & 64    & \cmark & 88    & +24 \\
    \bottomrule
    \end{tabular}%
  \label{tb:2}%
\end{table}%
Our results show that the models trained on synthetic data outperformed their counterparts only trained on the real data. The unfrozen FN+IN25 model achieved an 89\% accuracy on the real data when trained on synthetic data. In the frozen weights instances, the inception models perform similarly to the KNN models when trained only on the real data. However, the addition of synthetic data training improved the FZ model accuracy up to 83\% in the case of FN(FZ)+INC64. This addition also significantly impacted the I3D model, and its accuracy of 83\% outperforms all models not influenced by the synthetic data. Interestingly, this model even outperforms the FZ models trained on the synthetic data. This is quite impressive because I3D was designed for video action recognition tasks. Unlike the other models, the I3D had no prior information about the facial features.

In Table~\ref{tb:2} we explored the effect of unfreezing the FaceNet weights and the addition of synthetic data on the non-Caucasian data fold. We created one test fold which included 25 videos of the underrepresented ethnicities. The FN(FZ)+IN models trained without synthetic data are highlighted as the base models in this table. This table shows that the addition of synthetic data combined with unfreezing of the pre-trained weights has the highest impact on the correct classification of the non-Caucasian data samples (24\% increase). The addition of synthetic data alone has a limited beneficial impact; it can not alter the dataset bias effect of the original dataset on which the FaceNet was trained. 

\subsection{Combining synthetic and real data for training}
We designed another experiment to investigate how the accuracy changes with the addition of synthetic data. A portion of the synthetic data was randomly selected and combined with the real training data to create a new data set. The model was trained on this new training data and tested only on the real test data sample. We applied this training strategy to our most satisfactory model, input length 25 FaceNet + InceptionTime. In our first experiment, we set the ratio of selected synthetic data portion to 0.25. This ratio was doubled in the next experiment and doubled again in the last one. In each epoch $\left \lfloor ratio\times24 \right \rfloor$ human-like models were randomly selected. For every selected human-like model, out of the nine videos of that human-like model expressing a specific expression from multiple angles, only one was chosen randomly. This selection method means that only $\left \lfloor ratio\times24 \right \rfloor \times 21$ synthetic videos are used in the models training alongside the real data in that epoch. This number equals 126 for the ratio of 0.25, roughly equal to the number of real training videos. The selected human-like models and viewing angles were refreshed at the start of each epoch. 

\begin{figure}[t]

  \centering
  \includegraphics[width=0.7\textwidth]{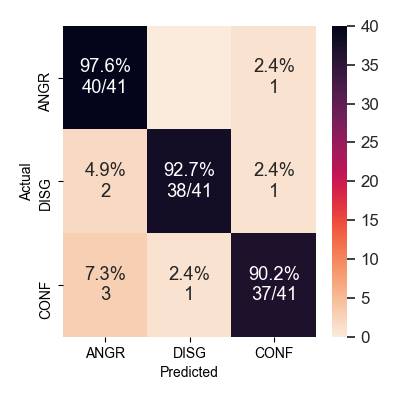}
  \caption{Combined confusion matrix of all the folds for the FN+INC64}\label{fig:cfm}
\end{figure}

The results for these experiments are shown in Table~\ref{tb:3}. These results show that doubling the synthetic data ratio from 0.5 to 1 increases the model's performance. However, this does not apply to the change from 0.25 to 0.5. In the case of FaceNet+Inception64, the synthetic ratio of 0.25 results in the highest performing network. This model achieved a 94\% accuracy, which shows an 18\% increase over the performance of the same model trained without the synthetic data. The combined confusion matrix of all the folds for this highest performing network is shown in Fig.~\ref{fig:cfm}.
\begin{table}[t]%FaceNet Inception time LEngth 25
\setlength{\tabcolsep}{2pt}
\renewcommand{\arraystretch}{0.01}
  \centering
  \caption{Performance comparison for the combined real and synthetic training method for the FN+INC models.}
  \begin{threeparttable}
    \begin{tabular}{cc||cccc}
\toprule
Ratio\tnote{*} & Length  & \%Precision   & \%Recall   & \%F-score    & \%Accuracy \\
\midrule
\midrule
0.25  & 25    & 89    & 87    & 87    & 87 \\
\midrule
0.25  & 64    & \textbf{95} & \textbf{94} & \textbf{94} & \textbf{94} \\
\midrule[1pt]

0.5   & 25    & 88    & 86    & 86    & 86 \\
\midrule
0.5   & 64    & 89    & 88    & 88    & 88 \\

\midrule[1pt]
1     & 25    & 89    & 87    & 87    & 88 \\
\midrule
1     & 64    & 89    & 88    & 88    & 88 \\
\bottomrule
\end{tabular}%
  
  \begin{tablenotes}
  %\item[*] Single  Image Input
  \item[*] The ratio for the selected proportion of the synthetic data. The 0.25 equals to 126 synthetic samples.
  \end{tablenotes}
  \end{threeparttable}\label{tb:3}%
\end{table}%

\subsection{Evaluating the Sim2Real approach on GIFGIF dataset}
\begin{table}[t]
  \centering
  \caption{Evaluation of Sim2Real effect on the model's performance on unseen data}
  \begin{threeparttable}
    \begin{tabular}{lcc||rrrr}
    \toprule
    \multicolumn{1}{c}{NET} & Len   & Synth\tnote{1} & \multicolumn{1}{c}{Prc\tnote{2}} & \multicolumn{1}{c}{Rec\tnote{3}} & \multicolumn{1}{c}{Fs\tnote{4}} & \multicolumn{1}{c}{Acc\tnote{5}} \\
    \midrule
    \midrule

    FN+INC25 & 25    & \xmark & 74    & 64    & 67    & 64 \\
    \midrule
    FN+INC25 & 25    & \cmark & 83    & 75    & 77    & 75 \\
    \bottomrule
    \end{tabular}%
   \begin{tablenotes}
   \item[1] Synthetic-Data \item[2]  Precision \item[3]  Recall \item[4]  F-score \item[5]  Accuracy
   \end{tablenotes}
  \end{threeparttable}\label{tb:unseen}
\end{table}%
% To evaluate the generalization of our Sim2Real approach and model, we selected an external dataset for cross-validation, GIFGIF~\cite{gifgif}.\footnote{We also considered AffWild, EmoReact, ElderReact, which either did not contain confusion and anger in their dataset, or did not have compatible annotation schemes (e.g., multi-label or frame-based).}
% As previously noted, there are currently no video datasets with confusion samples~\cite{Yasser2021}.
% GIFGIF~\cite{gifgif} has video-level annotations and contains 2 of our emotions of interest (``anger" and ``disgust").

% This dataset is a collection of 3,858
% cropped short videos with annotation scores for 17 emotions. We used GIFGIF API to get the first 400 highest-ranking videos for the ``disgust" emotion. These videos were filtered down to 75 based on the following criteria: 1) contains a human face reaction video, 2) must not hold a higher score in other categories. Similarly, we chose 75 ``anger" videos based on these criteria.
% We evaluated the two FN+IN25 models presented in Table~\ref{tb:1} ``as is", where one of these models had been pretrained on synthetic data. The GIFGIF dataset was used as a test dataset for these trained models. The results for this experiment is shown in Table~\ref{tb:unseen}.
To evaluate the generalization of our Sim2Real approach and model, we selected an external dataset for validation, GIFGIF~\cite{gifgif}.\footnote{We also considered AffWild, EmoReact, ElderReact, but their data did not contain anger or confusion, or their annotation schemes were not directly comparable with our data (e.g., frame-based). CK+~\cite{ck+}, Oulu-Casia~\cite{OUI}, and MMI~\cite{MMI} were also not selected since they are all acted/posed and we focus on in-the-wild interactions.} As previously noted, there are currently no video datasets with confusion samples~\cite{Yasser2021}. GIFGIF~\cite{gifgif} has video-level annotations and contains 2 of our emotions of interest (``anger" and ``disgust"). 

\begin{figure}[!t]
\centering
          \includegraphics[width=0.7\textwidth]{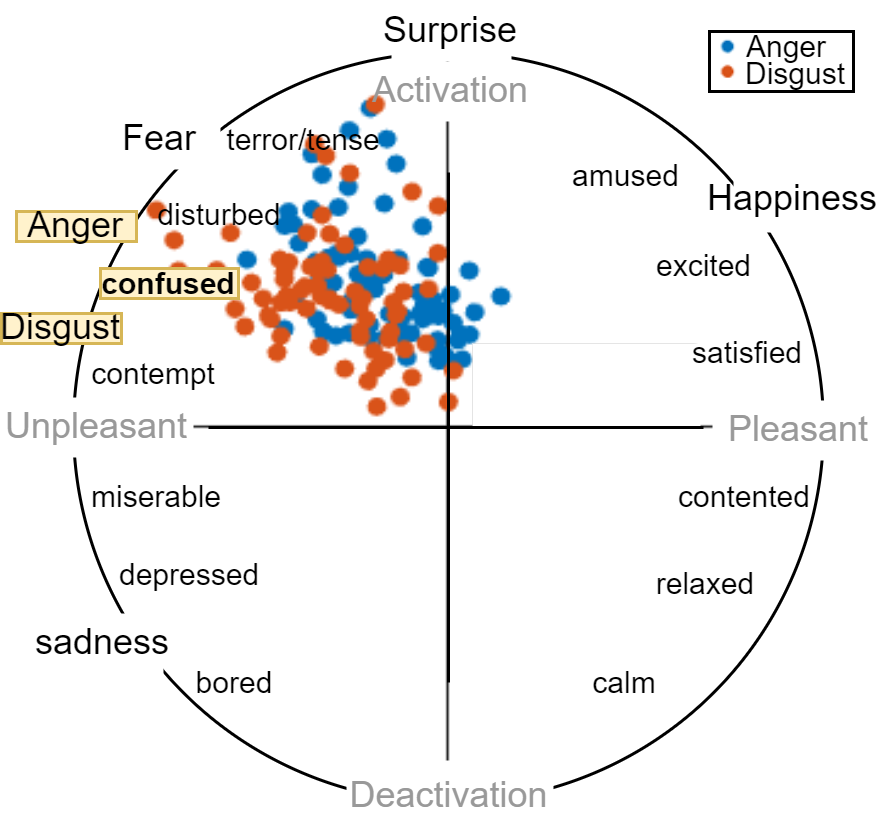}
         \caption{ Arousal-valence distribution of anger and disgust videos in GIFGIF~\cite{gifgif} using EmoNet~\cite{emonet} on CMA.}\label{fig:diag4}
     \hfill
\end{figure}

This dataset is a collection of 3,858 cropped short videos with annotation scores for 17 emotions. We used GIFGIF API to get the first 400 highest-ranking videos for the ``disgust" emotion. These videos were filtered down to 75 based on the following criteria: 1) contains a human face reaction video, 2) must not hold a higher score in other categories. Similarly, we chose the top 75 ``anger" videos. The Arousal-Valence distribution of all these 150 samples is displayed in Fig.~\ref{fig:diag4}. The Arousal-Valence values are extracted using Emonet~\cite{emonet}. Fig.~\ref{fig:diag4} suggests that this evaluation dataset is severely challenging.

For this evaluation, no additional training was performed. We tested the two FN+IN25 models presented in Table~\ref{tb:1} on this dataset. One model was trained on our real dataset, another model was pretrained on the synthetic dataset then trained on the real dataset. The GIFGIF dataset was used as a test dataset for these models. The results for this experiment are shown in Table~\ref{tb:unseen}. The FN+IN25 model pretrained on synthetic data achieved a 75\% accuracy. Out of 150 videos, this model misclassified 7 disgust videos and 2 anger as confusion. The same model without synthetic pretraining achieved a 64\% accuracy and misclassified 14 disgust and 4 anger videos as confusion. Therefore, without any additional transfer learning, we showed that our Sim2Real approach improved FN+INC25 performance on this dynamic FER task by 11\%.

% The FN+IN25 model pretrained on synthetic data achieved a 75\% accuracy. Out of 150 videos, this model misclassified 7 disgust videos and 2 anger as confusion. The same model without synthetic pretraining achieved a 64\% accuracy and misclassified 14 disgust and 4 anger videos as confusion. Therefore, without any additional transfer learning, we showed that our Sim2Real approach improved FN+INC25 performance on this dynamic FER task by 11\%.

\section{Discussion and Limitations}

In this section, we elaborate on insights that we found while doing experiments and after analyzing the results. Our experiments showed that additional synthetic data is similar to have an extensive dataset, and the generalization of the final model is increased.

An interesting finding in our experiments was that all the models with frozen FaceNet weights performed worse than their counterparts with unfrozen weights or even I3D. This was especially the case when considering non-Caucasian samples, which was an unexpected result because FaceNet was trained on a vast face recognition dataset. This shows that although the FaceNet feature embedding performs well on facial recognition tasks, it may not be entirely related to the facial changes of a specific emotional expression. However, more research is needed to investigate these hypotheses.
\begin{figure}[t]
\centering
          \includegraphics[width=0.7\textwidth]{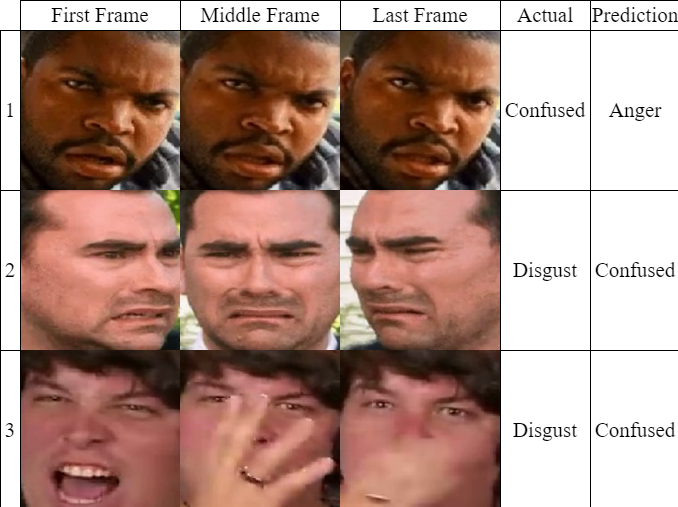}
         \caption{Sample of wrongly classified videos from our dataset}\label{fig:wr}
     \hfill
\end{figure}

Another interesting point was the misclassification of specific samples that were revealed after we looked deeper into the results. These samples were classified wrongly even in our best model with an average accuracy of 94\%,
Fig.~\ref{fig:wr} show three examples of the eight wrongly classified videos from all folds. From left to right, each column corresponds to the first, middle, and last frame. The incorrect classification of the first video might be related to the minimal movement of the face. The generated synthetic dataset that we used lacks fully static samples. Additionally, the main concept behind the proposed model was the focus on dynamic movements. The incorrect prediction of the second video relates to its head movement. The dynamic movement of the expression is done over frames involved with head movement. This adds fluctuation to the inception model's multivariant time-series input that may not relate to the emotion. The OpenFace algorithm~\cite{amos2016openface} uses perspective transform to make all of the input images have a frontal face view. The addition of this step may help deal with these types of videos. However, we believe the ultimate solution is in designing a model that can predict from shorter video snippets inputs. 
Poor prediction of the third video relates to the movement of hand midway through the video. Face occlusion is a challenge in FER, and even though new studies focus on reducing or removing the occlusions~\cite{Nojavanasghari2017,Naik2020}, their advancement has been minimal.

An interesting notion observed in the annotation of the real dataset was that annotators had trouble distinguishing between disgust and confusion in some cases. However, when the audio was played alongside the video, this confusion was resolved. This could mean that the next step for more inclusive and accurate facial expression recognition systems could incorporate audiovisual data processing.

One of the main components of this work was the generation of synthetic data. The MakeHuman application limitation highly affects this component. More advanced applications can be used for this task to generate more realistic synthetic datasets, and to explore other variations including age, non-binary gender, or conditions impacting facial development. Another point worth mentioning is that while we understand that the relatively small size of our dataset (synthetically generated dataset plus the real human dataset gathered from YouTube and Giphy) might be a limiting factor, this is sufficient to illustrate the proposed approach as a proof-of-concept.

\section{Conclusions and Future work}

We showed that our Sim2Real approach improves FN+INC64 performance on our dynamic FER task by 11-18\%, up to 94\% on our internal test dataset, and up to 75\% on a previously unseen dataset, compared to the performance of the same model architecture without synthetic training data. This performance was achieved by mixed synthetic and real data training. Additionally, it was shown that the proposed FN+INC model along with our Sim2Real approach is less sensitive to dataset ethnicity bias. This study was a first step towards emotion recognition in the wild, and future work can explore applying our approach and test the trained classifiers to data gathered from real-world HRI scenarios ~\cite{SahebJam2021}. Another notion that can be explored in the future is to observe the effects of replacing the face feature extractor model with a static FER feature extractor. An increased number of simulated human models may improve the overall accuracy, especially if they can be made with photo-realistic 3D model generating engines such as MetaHuman creator from Unreal Engine\footnote{\url{https://www.unrealengine.com}}. Automated animation generation from real videos can also be the next step for this study.

%%%%%%%%% REFERENCES
{\small
\bibliographystyle{IEEEtranS}
\bibliography{IEEEexample}
}

\end{document}